\documentclass[11pt]{article}
\usepackage{paclic34}
\usepackage{times}
\usepackage{latexsym}
\usepackage{amsmath}
\usepackage{multirow}
\usepackage{url}

\usepackage{multirow}
\usepackage{booktabs}

\usepackage{fontawesome}
\usepackage{float}
\usepackage{enumitem}
\usepackage{amsmath}
\usepackage{graphicx}
\newcommand{\sub}{\textsc{Slot-Sub}}
\newcommand{\sublm}{\textsc{Slot-Sub-LM}}
\newcommand{\crop}{\textsc{Crop}}
\newcommand{\rotate}{\textsc{Rotate}}
\usepackage{tikz-dependency}
\usepackage{subcaption}
\title{Simple is Better! Lightweight Data Augmentation for Low Resource Slot Filling and Intent Classification}

\author{Samuel Louvan \\
  University of Trento \\
  Fondazione Bruno Kessler  \\
  {\tt slouvan@fbk.eu} \\\And
  Bernardo Magnini \\
  Fondazione Bruno Kessler \\
  {\tt magnini@fbk.eu} \\}

\date{}

\begin{document}
\maketitle
\begin{abstract}
Neural-based models have achieved outstanding performance on slot filling and intent classification, when fairly large in-domain training data are available. However, as new domains are frequently added, creating sizeable data is expensive. We show that \textit{lightweight augmentation}, a set of augmentation methods involving word span and sentence level operations, alleviates data scarcity problems. Our experiments on limited data settings show that lightweight augmentation yields significant performance improvement on slot filling on the ATIS and SNIPS datasets, and achieves competitive performance with respect to more complex, state-of-the-art, augmentation approaches. Furthermore, lightweight augmentation is also beneficial when combined with pre-trained LM-based models, as it improves BERT-based joint intent and slot filling models.
\end{abstract}

\section{Introduction}

In task-oriented dialogue systems, a spoken language understanding component is responsible for parsing an utterance into a semantic representation. This is often modeled as a semantic frame \cite{tur2011spoken}, and typically involves \textit{slot filling} and \textit{intent classification}. For example, in the utterance "\textit{book in Southern Shores for 8 at Ariston Cafe}", the intent is \textit{booking a restaurant} and the corresponding \textit{slots} are Southern Shores (\textit{city-name}), 8 (\textit{number of people}), and Ariston Cafe (\textit{restaurant-name}).

Although neural-based models \cite{Qin2019ASF,goo2018slot,Mesnil2015UsingRN}  have achieved stellar performance in slot filling (SF) and intent classification (IC), their performance depend on the availability of  large labeled datasets. Consequently, they suffer in \textit{data scarcity} situations, which regularly happen when new domains are  added to the system to support new functionalities. 

One of the methods proposed to alleviate  data scarcity  is \textit{data augmentation} (DA), which aims to automatically increase the size of the training data by applying data transformations, ranging from simple word substitution to sentence generation. Recently, DA has shown promising potential for several NLP tasks, including text classification  \cite{DBLP:conf/emnlp/WeiZ19,DBLP:conf/emnlp/WangY15}, parsing \cite{DBLP:conf/emnlp/SahinS18,DBLP:conf/emnlp/VaniaKSL19}, and machine translation \cite{fadaee-etal-2017-data}. As for SF and IC, DA approaches typically generate synthetic utterances by leveraging Seq2Seq \cite{hou-etal-2018-sequence,DBLP:conf/emnlp/ZhaoZY19,DBLP:conf/emnlp/KurataXZY16}, Conditional VAE \cite{DBLP:conf/aaai/YooSL19}, or pre-trained NLG models \cite{DBLP:journals/corr/abs-2004-13952}. Such approaches make use of in-domain data, and  are relatively \textit{heavyweight}, as they require training neural models, which may involve several phases to generate, filter, and rank the produced augmented data, thus requiring more computation time.  It is also relatively challenging for deep learning-based models to generate semantically preserving synthetic utterances in limited data settings.
\begin{figure*}[htb!]
    \centering
    \includegraphics[width=.9\linewidth]{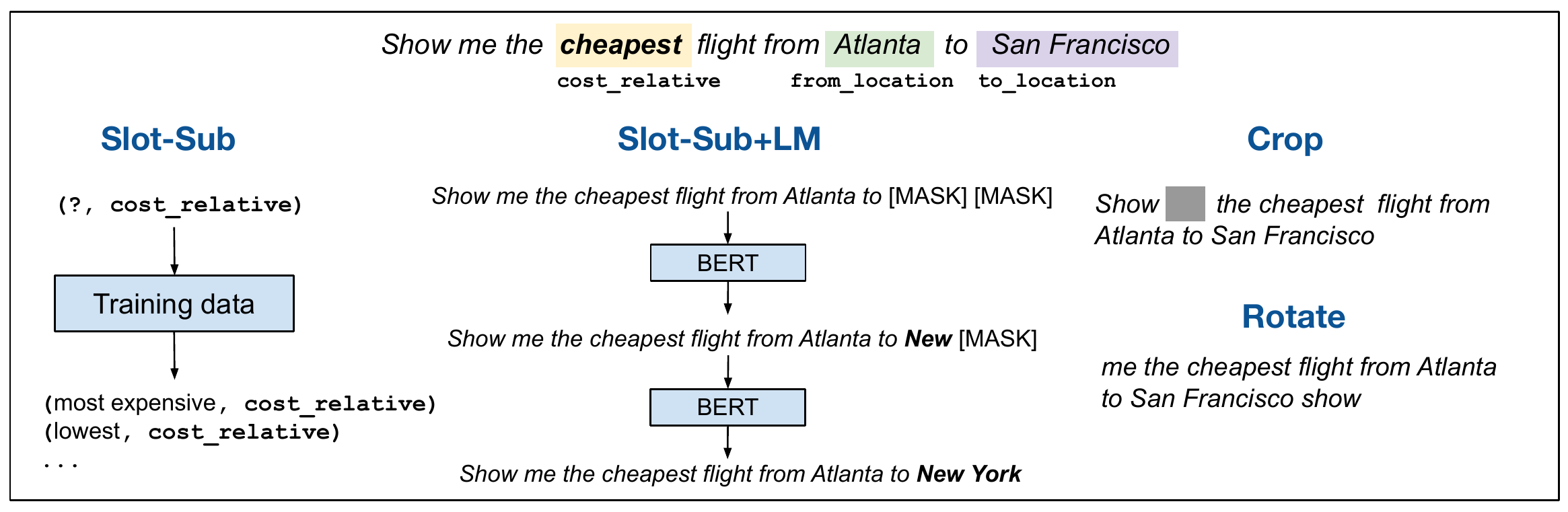}
    \caption{Examples of applying \textit{lightweight augmentation} on an utterance in the ATIS dataset.}
    \label{fig:augmentation}
\end{figure*}

In this paper, we show that \textit{lightweight augmentation}, a set of simple DA methods that produce utterance variations, is very effective for SF and IC in a low-resource setting. Lightweight augmentation considers both \textit{text span} and \textit{sentence} variations.  The span-level augmentation aims to diversify slot values in a particular text span through a \textit{semantically preserving} substitution of slot values. The sentence-level augmentation seeks to produce alternative sentence structure through crop and rotate \cite{DBLP:conf/emnlp/SahinS18} operations based on a dependency parse structure.

We investigate the effect of lightweight augmentation both on  typical biLSTM-based joint SF and IC models, and on large pre-trained LM transformers based models, in both cases with a limited data setting. Our contributions are as follows:
\begin{itemize}  
    \item We present a lightweight text span  and sentence level augmentation for SF and IC. We show that, despite its simplicity, lightweight augmentation is competitive with more complex, deep learning-based, augmentation. 
    \item We show that big self-supervised models, such as \textsc{BERT}\cite{Devlin2019BERTPO}, \textsc{RoBERTa}, and \textsc{ALBERT} can perform well under a low data regime, and still benefit from lightweight augmentation.
    \item The combination of our span based augmentation and transfer learning (e.g. \textsc{BERT} fine-tuning) yields the best performance for most cases. 
\end{itemize}

\section{Lightweight Data Augmentation}

Given the original training data $\mathcal{D}$, DA aims to generate additional training data $\mathcal{D}'$. For each sentence $S$ in $\mathcal{D}$, an augmentation operation is applied $N$ times, which can be empirically determined. Each augmented sentence $S'$ is added to $\mathcal{D}'$, and the union of $\mathcal{D}$ and $\mathcal{D}'$ is then used to train the model for SF and IC. We describe the lightweight DA operations in the following subsections.

\subsection{Slot Substitution (\sub)}
Our first lightweight method, slot substitution, is similar to \newcite{gulordava-etal-2018-colorless}, which is based on substituting a token in a sentence with another token with a consistent syntactic annotation (i.e., part-of-speech or morphology tags).  However, unlike \newcite{gulordava-etal-2018-colorless}, our method is not limited to single tokens. As slot filling is a \textit{semantic} task, rather than syntactic, we can naturally extend the method from single tokens (i.e., slot names composed by a single token) to  multiple tokens (i.e., slot names composed by multiple tokens, or \textit{spans}\footnote{We define a span as a sequence of one or more tokens that convey a slot value.}), still preserving the semantics associated to a certain slot.  

Pratically,  for slot substitution we take advantage of the fact that  SF training data are typically annotated with the BIO format\footnote{\texttt{B} indicates the beginning of the span, \texttt{I} indicates the inside of the span. \texttt{O} indicates that a token does not belong to any slot. For example, "San Francisco" will be annotated as \texttt{B-to\_location} \texttt{I-to\_location}.}.
We exploit the fact that two text spans in different utterances in $\mathcal{D}$ are likely to be semantically similar if they share the same slot label. We randomly pick one span in the $S$ and then perform the substitution (Figure \ref{fig:augmentation} \textit{Left}). 
For instance,  we can substitute the span "\textit{cheapest}", with other spans having the same slot label (i.e., \textsc{cost\_relative}), such as "\textit{lowest}" or "\textit{most expensive}". 

More formally, we denote a span $sp$ in a sentence $S$ as a slot-value pair $sp = (y, val)$, and we aim to produce an alternative pair $sp' = (y', val') $ such that the slot values are different  ($val \neq val'$) and the slot labels are the same ($y = y'$) for both slot-value pairs. To obtain $sp'$, we collect a set of candidates $\mathcal{SP}' = \{sp'_1, sp'_2,...,sp'_n \}$, by looking for slot spans in other sentences in $\mathcal{D}$ that satisfy our criteria. After that, we randomly sample a span from $\mathcal{SP}^{'}$ to obtain a $sp^{'}$. We replace $sp$ in $S$ with $sp'$ to produce the new augmented sentence $S^{'}$.  
For example, in the utterance "\textit{show me the cheapest flight from atlanta to san francisco}", one of the spans that can be substituted is $sp=(\textsc{cost\_relative}, "\mathit{cheapest}")$. Assuming that from $\mathcal{D}$ we can obtain $\mathcal{SP}^{aug}= \{ (\textsc{cost\_relative}, "\mathit{lowest}"), (\textsc{cost\_relative},\\ "\mathit{most\ expensive}"),\dots\}$, we then sample a $sp'$ from $\mathcal{SP}'$ and replace $sp$ in $S$ with $sp'$ to produce $S'$. Notice that the slot values in $sp'$ are not necessary synonyms of the original slot value, although their slot label must be the same to preserve  semantic compatibility.

\subsection{Slot Substitution with Language Model (\sublm)}
Our second lightweight method, \sublm, shares the goal with \sub, i.e., to substitute $sp$ with $sp'$. However, we do not use $\mathcal{D}$ to look for substitute candidates, instead we use a large pre-trained language model to generate the slot value candidates, using the \textit{fill-in-the-blank} style \cite{DBLP:conf/acl/DonahueLL20}. The expectation is that large pre-trained LMs, being trained on massive amount of data, can produce a sensible text span given a particular sentence context, and possibly produce slot values that do not occur in $\mathcal{D}$. While we use BERT for our purpose, virtually any pre-trained LM can be used for \sublm. Existing works on DA using LMs \cite{DBLP:conf/naacl/Kobayashi18,kumar2020data} are applied on text classification to replace random tokens in the text, which is not directly applicable to SF. Our approach focuses on \textit{spans}  conveying slot values, and include a filtering mechanism to reject retrieved slot spans that are not semantically compatible. 
\paragraph{Generating New Slot Values.} Given an utterance  consisting of one or more slot value spans, we "blank" one of the span and then let the LM to predict the new tokens in the span. For instance, we give
"\textit{show me the $\rule{1cm}{0.15mm}$ round trip flight from atlanta to denver}" to the LM for blank prediction. Practically, blank tokens are encoded as special \texttt{[MASK]} tokens\footnote{We set the number of masked tokens to be the same as the tokens of the original slot value, e.g. \texttt{san francisco} is masked as \texttt{[MASK][MASK]}, although this number could actually be sampled as well.} to let the pre-trained LM  performing prediction. The decoding of the new tokens is carried out iteratively from left to right (Figure \ref{fig:augmentation} \textit{Middle}) and, to produce the surface form of a token, we apply nucleus sampling \cite{DBLP:conf/iclr/HoltzmanBDFC20} using the top-\textit{p} portion of the probability mass. Nucleus sampling has been empirically shown to be better than beam search, and top-\textit{k} sampling \cite{DBLP:conf/acl/LewisDF18} to produce fluent and diverse texts.


\paragraph{Filtering.} 
While pre-trained LMs are expected to generate sensible replacements for a span in the utterance, a possible issue is that the new slot span is not semantically consistent with the original one. For example, for the original span "\textit{cheapest}" in  "\textit{show me the cheapest round trip flight from atlanta to denver}", the LM could output "\textit{earliest}" as a substitution, which does not fit the slot label \textsc{cost\_relative}. To mitigate this issue, we use a  binary sentence classifier as a \textit{filter} (\textbf{\sublm\texttt{+}Filter}) to decide whether $S$ and $S'$ are semantically compatible, based on the change made on the slot span. The training of the  classifier is composed of a pair  $S$ and $S'$, with its binary decision label (i.e., accept or reject $S'$). To construct the training data, for positive examples (\textit{accept}) we take advantage of the sentence pair produced by \sub, while for the negative examples (\textit{reject}) we sample $sp'$ in $\mathcal{D}$ where $y \neq  y'$ and replace $sp$ in $S$ with $sp'$ to produce $S'$. We use the BERT model as the sentence pair classifier and we encode the tokens, $w$, in both $S$ and $S'$ sentence pairs, as \texttt{[CLS]}$w^S_{1} w^S_{2}\dots w^{S}_n$ \texttt{[SEP]}$w^{S'}_1 w^{S'}_2\dots w^{S'}_m$. On top of BERT, we add a feed-forward layer that uses the sentence representation, \texttt{[CLS]}, for prediction. 

\subsection{\textsc{Crop} and \textsc{Rotate }}
\begin{figure}[t]
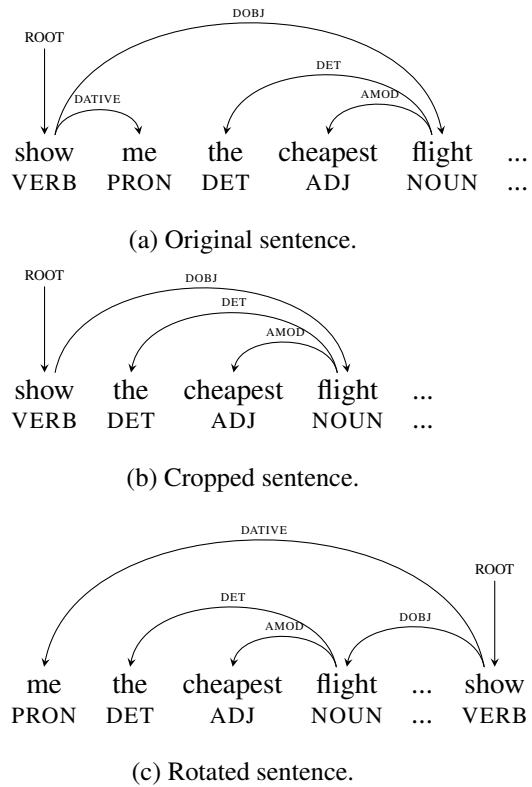

\centering
    \captionsetup[subfigure]{justification=centering}
    \begin{subfigure}[b]{0.4\textwidth}
    \begin{dependency}[theme=simple]
        \begin{deptext}[column sep=0.2cm]
            show \& me \& the \& cheapest \& flight \& ...\\
            \textsc{verb} \& \textsc{pron} \& \textsc{det} \& \textsc{adj} \& \textsc{noun} \& ... \\
        \end{deptext}
    \deproot[edge unit distance=2ex]{1}{ROOT}
    \depedge{1}{2}{\textsc{dative}}
    \depedge{1}{5}{\textsc{dobj}}
    \depedge{5}{3}{\textsc{det}}
    \depedge{5}{4}{\textsc{amod}}
    \end{dependency}
    \caption{Original sentence.}
    \label{fig:tm-orig-sent}
    \end{subfigure}
    
    \begin{subfigure}[b]{0.4\textwidth}
    \begin{dependency}[theme=simple]
        \begin{deptext}[column sep=0.2cm]
            show  \& the \& cheapest \& flight \& ...\\
            \textsc{verb}  \& \textsc{det} \& \textsc{adj} \& \textsc{noun} \& ... \\
        \end{deptext}
    \deproot[edge unit distance=2ex]{1}{ROOT}

    \depedge{1}{4}{\textsc{dobj}}
    \depedge{4}{2}{\textsc{det}}
    \depedge{4}{3}{\textsc{amod}}
    \end{dependency}
    \caption{Cropped sentence.}
    \label{fig:tm-cropped-sent}
    \end{subfigure}
    
    \begin{subfigure}[b]{0.4\textwidth}
    \begin{dependency}[theme=simple]
        \begin{deptext}[column sep=0.2cm]
            me  \& the \& cheapest \& flight \& ... \& show \\
            \textsc{pron}  \& \textsc{det} \& \textsc{adj} \& \textsc{noun} \& ... \& \textsc{verb} \\
        \end{deptext}
    \deproot[edge unit distance=2ex]{6}{ROOT}

    \depedge{6}{1}{\textsc{dative}}
    \depedge{6}{4}{\textsc{dobj}}
    \depedge{4}{3}{\textsc{amod}}
    \depedge{4}{2}{\textsc{det}}
    \end{dependency}
    \caption{Rotated sentence.}
    \label{fig:tm-rotated-sent}
    \end{subfigure}
\caption{Examples of dependency tree operations on a sentence.}
\label{fig:tree-morphing-example}
\end{figure}

The third lightweight method that we present augments an utterance by changing its syntactic structure. We adopt the augmentation approach from \cite{DBLP:conf/emnlp/SahinS18} (Figure \ref{fig:augmentation} \textit{Right}), which is based on two operations, \textit{crop} and \textit{rotate}, applied to the dependency parse tree of a  sentence. To our knowledge, this approach has not yet been applied to slot filling and intent classification, which is a contribution of our work.  \textit{Crop} focuses on particular fragments of a sentence (e.g., predicate and its subject, or predicate and its object), and removes the rest of the fragments, including its sub-tree, to create a smaller sentence.  \textit{Rotate}  aims to rotate the target fragment of a sentence  around the root of the dependency parse structure, producing a new utterance. For example, in the utterance "\textit{show me the cheapest flight from atlanta to san francisco}", the word \textit{"me"} can be cropped as it is one of the children of the \textit{root} verb "\textit{show}". While for rotation, the direct object (\textit{flight}) and its sub children (\textit{the cheapest}) are rotated around the root verb. Figure \ref{fig:tree-morphing-example} illustrates the relevant dependency structure manipulation. 
\section{Experiments and Results}
\begin{table*}[htb!]
\centering
    \small
    \begin{tabular}{lrr|rrr|cccc}
    \toprule
    & \multicolumn{2}{c|}{\textbf{Label}} & \multicolumn{3}{c|}{\textbf{\#Utterances} ($\mathcal{D}$)}  & \multicolumn{4}{c}{\#\textbf{Augmented Training Utterances} ($\mathcal{D'}$)} \\
    \cmidrule{2-10}
    \textbf{Dataset} & \#\textbf{slot} & \#\textbf{intent} & \#\textbf{train} &   \#\textbf{dev}  & \#\textbf{test} & \textbf{\sub} & \textbf{\sublm}  & \textsc{\textbf{\crop}} & \textsc{\textbf{\rotate}} \\
    \midrule
    ATIS & 79 & 18 & 0.4K & 500 & 893 & 3.9K & 0.8K&  0.8K& 1.1K \\
    SNIPS & 39 &  7 & 1.3K & 700 & 700 & 6.3K & 2.5K& 2.6K& 3.7K\\
    FB  & 16 & 12 & 3K & 4.1K& 8.6K & 5.4K & 5.4K & 5.9K &  8.5K\\
    \bottomrule
    \end{tabular}
    \caption{Statistics of both the original training data $\mathcal{D}$ and the augmented data $\mathcal{D'}$. \#train denotes our medium-size training data setup (10\% of full training data). $\mathcal{D'}$ is produced by each augmentation method, where the number $N$ of augmentations per sentence is tuned on the dev set.}
\label{tab:dataset}
\end{table*}
We experimented our lightweight augmentation approach on three well-known  datasets for SF and IC, namely ATIS \cite{Hemphill1990TheAS}, SNIPS \cite{Coucke2018SnipsVP} and FB \cite{Schuster2018CrosslingualTL}. All datasets are in English. ATIS contains utterances related to flight domain (e.g., searching flight, booking). SNIPS includes multi-domain utterances such as weather, movie, restaurant, etc. FB contains utterances from 3 domains, weather, alarm, and reminder. To simulate the \textit{data scarcity} setting, we follow previous works \cite{hou-etal-2018-sequence,DBLP:conf/aaai/YooSL19} and only use \textit{medium}-size (i.e., 1/10) of training data for each dataset. Statistics on the three datasets are reported in Table \ref{tab:dataset}.

As for evaluation, we use standard evaluation metrics, namely the F1-score for SF and accuracy for IC\footnote{Metric is computed using the standard evaluation script https://www.clips.uantwerpen.be/conll2000/}. Performance are calculated as the average score of ten different runs. In order to compare our methods, we use two  baselines for slot filling and intent detection: a simple BiLSTM-CRF model, and a state of the art BERT-based model, which is fine-tuned to SF and IC\footnote{We use the \texttt{bert-base-uncased} model}. Each model is trained for 30 epochs, and we apply early stopping criteria.

For the slot substitution (\sub) and the slot substitution with language model (\sublm) augmentation methods, we tune the number of augmented sentence per utterance, $N$, on the dev set of each dataset. For crop and rotate,  we use the default parameters from \newcite{DBLP:conf/emnlp/SahinS18}. To produce the dependency parse structure for the utterances in our datasets, we use Spacy\footnote{https://spacy.io/}. All hyperparameters are tuned on the dev set.  More details on the settings is provided in Appendix A. 

In order to allow comparison with more complex data augmentation approaches, we also report results obtained with state of the art approaches based on Seq2Seq \cite{hou-etal-2018-sequence} and Conditional Variational Auto Encoder (CVAE) \cite{DBLP:conf/aaai/YooSL19}. Our implementation is based on the Huggingface library \cite{wolf2019transformers}, and will be made publicly available.

\subsection{Results}

\begin{table*}[!h]
    \centering
    \begin{tabular}{l l c c c c c c c c}
    \toprule
    \multirow{2}{*}{\textbf{Model}} & \multirow{2}{*}{\textbf{DA}} & \multicolumn{2}{c}{\textbf{ATIS}} & \multicolumn{2}{c}{\textbf{SNIPS}}  & \multicolumn{2}{c}{\textbf{FB}}\\
    \cmidrule{3-8}
    & & Slot & Intent  & Slot & Intent & Slot & Intent  \\
    \midrule
    BiLSTM+CRF  & None  &  86.83 & 90.64 & 84.51 & 95.94 & 93.83 & 98.47 \\
                     & Seq2Seq \cite{DBLP:conf/coling/HouLCL18}&  88.72 & - & - & - & - & -  \\
                    & VAE \cite{DBLP:conf/aaai/YooSL19}  & 89.27 & 90.95 & - & - & - & - \\
                    \cmidrule{2-8}
                    & \sub& $\underline{89.89}^{\dagger}$	& $\underline{93.37}^{\dagger}$	 & $\underline{86.45}^{\dagger}$ & $96.30^{\dagger}$ &	93.70 & 98.45\\
                    & \sublm& 87.03	& $92.96^{\dagger}$&	82.82 & 96.14 &91.52&98.20\\
                     & \sublm +Filter&  87.19 &	$92.01^{\dagger}$ &	                    82.77	& 96.08	& 92.18	& 98.37\\
                    & \textsc{Crop}& $88.62^{\dagger}$ &	$92.32^{\dagger}$ &	$85.84^{\dagger}$ &	96.07 &	93.91 &	\underline{98.64}\\
                    & \textsc{Rotate}& $88.83^{\dagger}$ &	$92.33^{\dagger}$ &	85.65 &	$\underline{96.39}^{\dagger}$ &	$\underline{94.04}$ & 98.56\\
              
    \midrule
    BERT   & None & 89.39	& 94.98 &	89.17 &	96.70&	94.22&	98.61\\
                    \cmidrule{2-8}
                    & \sub & $\textbf{\underline{90.43}}^{\dagger}$	& $\textbf{\underline{95.49}}^{\dagger}$ &	$\textbf{\underline{90.66}}^{\dagger}$ &$	\textbf{\underline{97.11}}^{\dagger} $&	94.01 &	98.59\\
                    & \textsc{Crop} & 89.47	& 94.55 &	89.77 &	96.78 &	94.20 &	98.73 \\
                    & \textsc{Rotate} & 89.57 &	94.48 &	89.37 &	96.81 &	\textbf{\underline{94.32}} &	\textbf{\underline{98.80}} \\
    \bottomrule
    \end{tabular}
    \caption{Overall results on the test set.  Underlined numbers indicate best performing methods for a particular slot filling + intent model.  \textbf{Bold} numbers indicate  best overall methods. $\dagger$ indicates significant improvement over the baseline without augmentation ( \textit{p}-value $<$ 0.05, Wilcoxon signed rank test). We do not apply \sublm\  to the BERT slot filling and intent model because we also use BERT for \sublm, so we think this is redundant.}
    \label{tab:overall}
\end{table*}

Table \ref{tab:overall} reports the results on the \textit{test sets} used in our experiments. We include best-reported scores from two state of the art augmentation methods for comparison, namely a sequence-to-sequence (Seq2Seq) based from \newcite{hou-etal-2018-sequence} and a VAE based methods from \newcite{DBLP:conf/aaai/YooSL19}. Results in Table \ref{tab:overall} (\textit{test set}) show that lightweight augmentation is beneficial for both Bi-LSTM CRF and BERT, on both ATIS (single domain) and SNIPS (multi-domain) datasets.  \sub\ yields the best results for both the BiLSTM+CRF and BERT models, with SF performance up to 90.43 on ATIS and 90.66 on SNIPS, and IC performance to 95.49 on ATIS and 97.11 on SNIPS. As for the FB dataset, models only gain marginal improvement across lightweight augmentation. We hypothesize that FB  is relatively easy to solve, compared with ATIS and SNIPS, as the slot filling performance of BiLSTM without augmentation already achieves a very high F1 score. The improvement using augmentation is more significant for SF rather than IC.


Out of all lightweight augmentation methods,   \sub\ obtains the best performance, particularly on slot filling on ATIS and SNIPS. The overall best performing configuration is a combination of BERT fine-tuning with  \sub\ augmentation. Given  limited training data, BERT fine-tuning without augmentation surpasses BiLSTM-CRF without augmentation by a large margin. Yet, performance can be boosted even further with lightweight augmentation, suggesting that even a big, self-supervised model, such as BERT can still benefit from  augmentation on limited data settings. The improvements on BiLSTM-CRF indicate that lightweight augmentation  improves the model's robustness when trained on small amounts of data. We find that \sublm\ is suboptimal for SF. Our qualitative observation shows that \sublm\ often generates slot values that are semantically incompatible with the original slot label.  \textsc{Crop} and \textsc{Rotate} can help IC in some cases although their improvement is marginal.  

Despite its simplicity, \sub\ is also competitive with state-of-the-art heavyweight data augmentation approaches (Seq2Seq and CVAE), significantly boosting Bi-LSTM and BERT performance for SF on ATIS and SNIPS. We believe that the key advantage of \sub\ is its capability to maintain semantic consistency over the slot spans, which has revealed to be  stronger than that of heavyweight approaches. This also shows that slot consistency is crucial for obtaining good performance, particularly for SF.  While the CVAE based approach from \newcite{DBLP:conf/aaai/YooSL19} has injected slot and intent labels in the model, it seems that generating semantically consistent utterance is still challenging for deep learning models, especially when data is limited.

\section{Analysis and Discussion}
In this Section we discuss several aspects of data augmentation applied to slot filling and intent detection.

\paragraph{Impact of number of augmented sentences.}

To better understand the effect of the number of augmented sentences per utterance ($N$), we now observe the performance of our best performing method, \sub, while changing $N$ values (we use $\{2, 5, 10, 20, 25\}$) on the dev set. As for ATIS, increasing $N$ yields a F1  improvement from 90.68 up to 91.62; SNIPS performance increased from 87 F1 and to 88 F1 when increasing $N$ from 2 to 5 and it is stable around 88 F1 when using $N$ larger than 5; finally, FB is stable around 93.4 to 93.7 F1. Overall, the biggest improvement is when $N$ is increased from 2 to 5, while with higher values only minor improvements can still be obtained on ATIS. 

\paragraph{Performance on different training data size ($\mathcal{D}$).} 
\begin{figure}[!ht]
    \includegraphics[width=\linewidth]{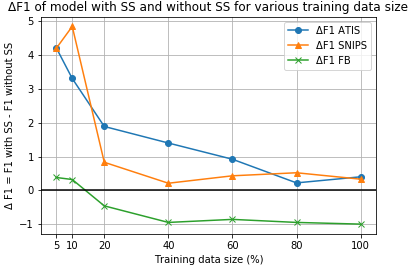}
    \caption{Gain ($\Delta F1$) obtained by \sub\ (SS) on various training data size. Positive numbers mean that the model with SS is better than without SS.} 
    
    \label{fig:increase}
\end{figure}

Figure \ref{fig:increase} displays the gain obtained by \sub\ for various data size for slot filling. Using smaller data size (i.e., 5\%) than our default setting, \sub\ still obtains a F1 gain for all datasets. On the other hand, as we increase the number of training data, the \sub\ benefit  diminishes, without hurting performance on ATIS and SNIPS. As for FB we observe a performance drop of less than 1 F1, which is still relatively low. 

\paragraph{Is lightweight augmentation beneficial to very large language models?} 
\begin{table}
    \small
    \centering
    \begin{tabular}{l l c c c c c c}
    \toprule
    \multirow{2}{*}{\textbf{Model}} & \multirow{2}{*}{\textbf{Aug.}} & \multicolumn{2}{c}{\textbf{ATIS}} & \multicolumn{2}{c}{\textbf{SNIPS}}  \\
    \cmidrule{3-6}
    & & Slot & Intent  & Slot & Intent  \\
    \midrule
    BERT   & None  & 91.6	& 95.0 & 89.8	& 95.0 \\
            
    (large)             & \textsc{SS} & \underline{92.8} & \underline{95.4} & \underline{92.8}	& \underline{95.4}\\
    \midrule
    Albert   & None  & 92.1& 94.8 & 89.5 & 99.0 \\
    (xxl)                    & \textsc{SS} & \underline{92.9}	& \underline{95.0}  & \underline{\textbf{93.6}} & \underline{\textbf{99.2}} \\
    \midrule
    Roberta   & None  & 90.6	& 92.8 & 89.2	& \underline{98.9}\\
    (large)                & \textsc{SS} & \underline{\textbf{93.2}} & \textbf{\underline{95.9}} & \underline{92.5} &	98.8 \\
    \bottomrule
    \end{tabular}
    \caption{Lightweight augmentation \sub\ (SS) applied to very large pre-trained LMs.}
    \label{tab:compare_large_models}
\end{table}
Motivated by the encouraging results that lightweight augmentation has obtained on a strong pre-trained LM such as BERT on low-resource settings (see Table \ref{tab:overall}),  we now further examine the advantage of lightweight augmentation for other very large pre-trained LM models, namely Albert \cite{DBLP:conf/iclr/LanCGGSS20} and Roberta \cite{liu2019roberta}. We use the largest trained models for each of the pre-trained LM, namely \texttt{bert-large-uncased}, \texttt{roberta-large}, and \texttt{albert-xxl}. Results, reported in  Table \ref{tab:compare_large_models}, show that on limited data settings, all the very large models still benefit from \sub, notably on the performance for SF.


\section{Related Work}

Data augmentation methods have been widely applied in computer vision, ranging from geometric transformations \cite{krizhevsky2012imagenet,zhong2020random}, data mixing \cite{summers2019improved} to the use of generative models \cite{goodfellow2014generative} for generating synthetic data. Recently, data augmentation has been applied to various NLP tasks, including text classification  \cite{DBLP:conf/emnlp/WeiZ19,DBLP:conf/emnlp/WangY15}, parsing \cite{DBLP:conf/emnlp/SahinS18,DBLP:conf/emnlp/VaniaKSL19}, and machine translation \cite{fadaee-etal-2017-data}. Augmentation techniques for NLP tasks range from operations on tokens (e.g., substituting, deleting) \cite{DBLP:conf/emnlp/WangY15,DBLP:conf/naacl/Kobayashi18,DBLP:conf/emnlp/WeiZ19}, to manipulation of the sentence structure  \cite{sahin-steedman-2018-data}, to paraphrase-based augmentation \cite{callison-burch-etal-2006-improved}. 

Data augmentation has been also experimented in the context of slot filling and intent classification. Particularly, recent methods have focused on the application of generative models to produce synthetic utterances. \newcite{DBLP:conf/coling/HouLCL18} proposes a method that separates the utterance generation from the slot values realization. A sequence to sequence based model is used to generate utterances for a given intent with slot values placeholders (i.e., delexicalized), and then words in the training data that occur in similar context of the placeholder are inserted as the slot values. 
\newcite{zhao-etal-2019-data} also uses a sequence to sequence model by exploiting a small number of template exemplars. \newcite{DBLP:conf/aaai/YooSL19} proposes a solution based on Conditional Variational Auto Encoder (CVAE) to generate synthetic utterances. In this case the CVAE takes into account both the intent and the slot labels during training, and the model generates the surface form of the utterance, slot labels, and the intent label. Recent work from \newcite{peng2020data} make use of GPT-2 \cite{radford2019language}, and fine-tuned it to intent and slot-value pairs to generate utterances. 

In comparison to existing, state of the art, augmentation methods for slot filling and intent detection, the augmentation methods proposed in this paper can be considered as \textit{lightweight} because they do not require any separate training based on deep learning models for generating additional data. 
Still, lightweight augmentation maintains consistent slot semantic substitutions, a feature that is crucial for effective data augmentation. In the spectrum of existing augmentation methods, i.e., from words manipulation to paraphrasing-based methods, our lightweight approaches lie in the middle, as we focus on particular \textit{text spans} that convey slot values or on particular structures in the dependency parse tree of the utterance.


\section{Conclusion}
We showed that lightweight augmentation for slot filling and and intent detection in low-resource settings is very competitive with respect to more complex deep learning based data augmentation. A lightweight method based on slot values substitution, while preserving the semantic consistency of slot labels,  has proven to be the more effective.  We also show that large self-supervised models like BERT can benefit from lightweight augmentation, suggesting that a \textit{combination} of data augmentation and transfer learning is very useful, and has the potential to be applied to other NLP tasks. 

For future work, it would be interesting to see the effect of using the augmented data generated by \sub\ as additional training data for deep learning based augmentation models. Encouraged by the results of our lightweight augmentation, our work can also be experimented on semantic tasks with similar characteristics, such as Named Entity Recognition.

\bibliography{anthology, paclic}

\begin{thebibliography}{}

\bibitem[\protect\citename{Callison-Burch \bgroup et al.\egroup
  }2006]{callison-burch-etal-2006-improved}
Chris Callison-Burch, Philipp Koehn, and Miles Osborne.
\newblock 2006.
\newblock Improved statistical machine translation using paraphrases.
\newblock In {\em Proceedings of the Human Language Technology Conference of
  the {NAACL}, Main Conference}, pages 17--24, New York City, USA, June.
  Association for Computational Linguistics.

\bibitem[\protect\citename{Coucke \bgroup et al.\egroup
  }2018]{Coucke2018SnipsVP}
Alice Coucke, Alaa Saade, Adrien Ball, Th{\'e}odore Bluche, Alexandre Caulier,
  David Leroy, Cl{\'e}ment Doumouro, Thibault Gisselbrecht, Francesco
  Caltagirone, Thibaut Lavril, Ma{\"e}l Primet, and Joseph Dureau.
\newblock 2018.
\newblock Snips voice platform: an embedded spoken language understanding
  system for private-by-design voice interfaces.
\newblock {\em ArXiv}, abs/1805.10190.

\bibitem[\protect\citename{Devlin \bgroup et al.\egroup
  }2019]{Devlin2019BERTPO}
Jacob Devlin, Ming-Wei Chang, Kenton Lee, and Kristina Toutanova.
\newblock 2019.
\newblock Bert: Pre-training of deep bidirectional transformers for language
  understanding.
\newblock In {\em NAACL-HLT}.

\bibitem[\protect\citename{Donahue \bgroup et al.\egroup
  }2020]{DBLP:conf/acl/DonahueLL20}
Chris Donahue, Mina Lee, and Percy Liang.
\newblock 2020.
\newblock Enabling language models to fill in the blanks.
\newblock In Dan Jurafsky, Joyce Chai, Natalie Schluter, and Joel~R. Tetreault,
  editors, {\em Proceedings of the 58th Annual Meeting of the Association for
  Computational Linguistics, {ACL} 2020, Online, July 5-10, 2020}, pages
  2492--2501. Association for Computational Linguistics.

\bibitem[\protect\citename{Fadaee \bgroup et al.\egroup
  }2017]{fadaee-etal-2017-data}
Marzieh Fadaee, Arianna Bisazza, and Christof Monz.
\newblock 2017.
\newblock Data augmentation for low-resource neural machine translation.
\newblock In {\em Proceedings of the 55th Annual Meeting of the Association for
  Computational Linguistics (Volume 2: Short Papers)}, pages 567--573,
  Vancouver, Canada, July. Association for Computational Linguistics.

\bibitem[\protect\citename{Fan \bgroup et al.\egroup
  }2018]{DBLP:conf/acl/LewisDF18}
Angela Fan, Mike Lewis, and Yann~N. Dauphin.
\newblock 2018.
\newblock Hierarchical neural story generation.
\newblock In Iryna Gurevych and Yusuke Miyao, editors, {\em Proceedings of the
  56th Annual Meeting of the Association for Computational Linguistics, {ACL}
  2018, Melbourne, Australia, July 15-20, 2018, Volume 1: Long Papers}, pages
  889--898. Association for Computational Linguistics.

\bibitem[\protect\citename{Goo \bgroup et al.\egroup }2018]{goo2018slot}
Chih-Wen Goo, Guang Gao, Yun-Kai Hsu, Chih-Li Huo, Tsung-Chieh Chen, Keng-Wei
  Hsu, and Yun-Nung Chen.
\newblock 2018.
\newblock Slot-gated modeling for joint slot filling and intent prediction.
\newblock In {\em Proceedings of the 2018 Conference of the North American
  Chapter of the Association for Computational Linguistics: Human Language
  Technologies, Volume 2 (Short Papers)}, pages 753--757.

\bibitem[\protect\citename{Goodfellow \bgroup et al.\egroup
  }2014]{goodfellow2014generative}
Ian Goodfellow, Jean Pouget-Abadie, Mehdi Mirza, Bing Xu, David Warde-Farley,
  Sherjil Ozair, Aaron Courville, and Yoshua Bengio.
\newblock 2014.
\newblock Generative adversarial nets.
\newblock In {\em Advances in neural information processing systems}, pages
  2672--2680.

\bibitem[\protect\citename{Gulordava \bgroup et al.\egroup
  }2018]{gulordava-etal-2018-colorless}
Kristina Gulordava, Piotr Bojanowski, Edouard Grave, Tal Linzen, and Marco
  Baroni.
\newblock 2018.
\newblock Colorless green recurrent networks dream hierarchically.
\newblock In {\em Proceedings of the 2018 Conference of the North {A}merican
  Chapter of the Association for Computational Linguistics: Human Language
  Technologies, Volume 1 (Long Papers)}, pages 1195--1205, New Orleans,
  Louisiana, June. Association for Computational Linguistics.

\bibitem[\protect\citename{Hemphill \bgroup et al.\egroup
  }1990]{Hemphill1990TheAS}
Charles~T. Hemphill, John~J. Godfrey, and George~R. Doddington.
\newblock 1990.
\newblock The atis spoken language systems pilot corpus.
\newblock In {\em HLT}.

\bibitem[\protect\citename{Holtzman \bgroup et al.\egroup
  }2020]{DBLP:conf/iclr/HoltzmanBDFC20}
Ari Holtzman, Jan Buys, Li~Du, Maxwell Forbes, and Yejin Choi.
\newblock 2020.
\newblock The curious case of neural text degeneration.
\newblock In {\em 8th International Conference on Learning Representations,
  {ICLR} 2020, Addis Ababa, Ethiopia, April 26-30, 2020}. OpenReview.net.

\bibitem[\protect\citename{Hou \bgroup et al.\egroup
  }2018a]{hou-etal-2018-sequence}
Yutai Hou, Yijia Liu, Wanxiang Che, and Ting Liu.
\newblock 2018a.
\newblock Sequence-to-sequence data augmentation for dialogue language
  understanding.
\newblock In {\em Proceedings of the 27th International Conference on
  Computational Linguistics}, pages 1234--1245, Santa Fe, New Mexico, USA,
  August. Association for Computational Linguistics.

\bibitem[\protect\citename{Hou \bgroup et al.\egroup
  }2018b]{DBLP:conf/coling/HouLCL18}
Yutai Hou, Yijia Liu, Wanxiang Che, and Ting Liu.
\newblock 2018b.
\newblock Sequence-to-sequence data augmentation for dialogue language
  understanding.
\newblock In Emily~M. Bender, Leon Derczynski, and Pierre Isabelle, editors,
  {\em Proceedings of the 27th International Conference on Computational
  Linguistics, {COLING} 2018, Santa Fe, New Mexico, USA, August 20-26, 2018},
  pages 1234--1245. Association for Computational Linguistics.

\bibitem[\protect\citename{Kobayashi}2018]{DBLP:conf/naacl/Kobayashi18}
Sosuke Kobayashi.
\newblock 2018.
\newblock Contextual augmentation: Data augmentation by words with paradigmatic
  relations.
\newblock In Marilyn~A. Walker, Heng Ji, and Amanda Stent, editors, {\em
  Proceedings of the 2018 Conference of the North American Chapter of the
  Association for Computational Linguistics: Human Language Technologies,
  NAACL-HLT, New Orleans, Louisiana, USA, June 1-6, 2018, Volume 2 (Short
  Papers)}, pages 452--457. Association for Computational Linguistics.

\bibitem[\protect\citename{Krizhevsky \bgroup et al.\egroup
  }2012]{krizhevsky2012imagenet}
Alex Krizhevsky, Ilya Sutskever, and Geoffrey~E Hinton.
\newblock 2012.
\newblock Imagenet classification with deep convolutional neural networks.
\newblock In {\em Advances in neural information processing systems}, pages
  1097--1105.

\bibitem[\protect\citename{Kumar \bgroup et al.\egroup }2020]{kumar2020data}
Varun Kumar, Ashutosh Choudhary, and Eunah Cho.
\newblock 2020.
\newblock Data augmentation using pre-trained transformer models.
\newblock {\em arXiv preprint arXiv:2003.02245}.

\bibitem[\protect\citename{Kurata \bgroup et al.\egroup
  }2016]{DBLP:conf/emnlp/KurataXZY16}
Gakuto Kurata, Bing Xiang, Bowen Zhou, and Mo~Yu.
\newblock 2016.
\newblock Leveraging sentence-level information with encoder {LSTM} for
  semantic slot filling.
\newblock In Jian Su, Xavier Carreras, and Kevin Duh, editors, {\em Proceedings
  of the 2016 Conference on Empirical Methods in Natural Language Processing,
  {EMNLP} 2016, Austin, Texas, USA, November 1-4, 2016}, pages 2077--2083. The
  Association for Computational Linguistics.

\bibitem[\protect\citename{Lan \bgroup et al.\egroup
  }2020]{DBLP:conf/iclr/LanCGGSS20}
Zhenzhong Lan, Mingda Chen, Sebastian Goodman, Kevin Gimpel, Piyush Sharma, and
  Radu Soricut.
\newblock 2020.
\newblock {ALBERT:} {A} lite {BERT} for self-supervised learning of language
  representations.
\newblock In {\em 8th International Conference on Learning Representations,
  {ICLR} 2020, Addis Ababa, Ethiopia, April 26-30, 2020}. OpenReview.net.

\bibitem[\protect\citename{Liu \bgroup et al.\egroup }2019]{liu2019roberta}
Yinhan Liu, Myle Ott, Naman Goyal, Jingfei Du, Mandar Joshi, Danqi Chen, Omer
  Levy, Mike Lewis, Luke Zettlemoyer, and Veselin Stoyanov.
\newblock 2019.
\newblock Roberta: A robustly optimized bert pretraining approach.
\newblock {\em arXiv preprint arXiv:1907.11692}.

\bibitem[\protect\citename{Mesnil \bgroup et al.\egroup
  }2015]{Mesnil2015UsingRN}
Gr{\'e}goire Mesnil, Yann Dauphin, Kaisheng Yao, Yoshua Bengio, Li~Deng,
  Dilek~Z. Hakkani-Tur, Xiaodong He, Larry Heck, Gokhan Tur, Dong Yu, and
  Geoffrey Zweig.
\newblock 2015.
\newblock Using recurrent neural networks for slot filling in spoken language
  understanding.
\newblock {\em IEEE/ACM Transactions on Audio, Speech, and Language
  Processing}, 23:530--539.

\bibitem[\protect\citename{Peng \bgroup et al.\egroup
  }2020a]{DBLP:journals/corr/abs-2004-13952}
Baolin Peng, Chenguang Zhu, Michael Zeng, and Jianfeng Gao.
\newblock 2020a.
\newblock Data augmentation for spoken language understanding via pretrained
  models.
\newblock {\em CoRR}, abs/2004.13952.

\bibitem[\protect\citename{Peng \bgroup et al.\egroup }2020b]{peng2020data}
Baolin Peng, Chenguang Zhu, Michael Zeng, and Jianfeng Gao.
\newblock 2020b.
\newblock Data augmentation for spoken language understanding via pretrained
  models.
\newblock {\em arXiv preprint arXiv:2004.13952}.

\bibitem[\protect\citename{Qin \bgroup et al.\egroup }2019]{Qin2019ASF}
Libo Qin, Wanxiang Che, Yangming Li, Haoyang Wen, and Ting Liu.
\newblock 2019.
\newblock A stack-propagation framework with token-level intent detection for
  spoken language understanding.
\newblock In {\em EMNLP/IJCNLP}.

\bibitem[\protect\citename{Radford \bgroup et al.\egroup
  }2019]{radford2019language}
Alec Radford, Jeffrey Wu, Rewon Child, David Luan, Dario Amodei, and Ilya
  Sutskever.
\newblock 2019.
\newblock Language models are unsupervised multitask learners.

\bibitem[\protect\citename{Sahin and Steedman}2018a]{DBLP:conf/emnlp/SahinS18}
G{\"{o}}zde~G{\"{u}}l Sahin and Mark Steedman.
\newblock 2018a.
\newblock Data augmentation via dependency tree morphing for low-resource
  languages.
\newblock In Ellen Riloff, David Chiang, Julia Hockenmaier, and Jun'ichi
  Tsujii, editors, {\em Proceedings of the 2018 Conference on Empirical Methods
  in Natural Language Processing, Brussels, Belgium, October 31 - November 4,
  2018}, pages 5004--5009. Association for Computational Linguistics.

\bibitem[\protect\citename{{\c{S}}ahin and
  Steedman}2018b]{sahin-steedman-2018-data}
G{\"o}zde~G{\"u}l {\c{S}}ahin and Mark Steedman.
\newblock 2018b.
\newblock Data augmentation via dependency tree morphing for low-resource
  languages.
\newblock In {\em Proceedings of the 2018 Conference on Empirical Methods in
  Natural Language Processing}, pages 5004--5009, Brussels, Belgium,
  October-November. Association for Computational Linguistics.

\bibitem[\protect\citename{Schuster \bgroup et al.\egroup
  }2018]{Schuster2018CrosslingualTL}
Sebastian Schuster, Sonal Gupta, Rushin Shah, and Mike Lewis.
\newblock 2018.
\newblock Cross-lingual transfer learning for multilingual task oriented
  dialog.
\newblock In {\em NAACL-HLT}.

\bibitem[\protect\citename{Summers and Dinneen}2019]{summers2019improved}
Cecilia Summers and Michael~J Dinneen.
\newblock 2019.
\newblock Improved mixed-example data augmentation.
\newblock In {\em 2019 IEEE Winter Conference on Applications of Computer
  Vision (WACV)}, pages 1262--1270. IEEE.

\bibitem[\protect\citename{Tur and De~Mori}2011]{tur2011spoken}
Gokhan Tur and Renato De~Mori.
\newblock 2011.
\newblock {\em Spoken language understanding: Systems for extracting semantic
  information from speech}.
\newblock John Wiley \& Sons.

\bibitem[\protect\citename{Vania \bgroup et al.\egroup
  }2019]{DBLP:conf/emnlp/VaniaKSL19}
Clara Vania, Yova Kementchedjhieva, Anders S{\o}gaard, and Adam Lopez.
\newblock 2019.
\newblock A systematic comparison of methods for low-resource dependency
  parsing on genuinely low-resource languages.
\newblock In Kentaro Inui, Jing Jiang, Vincent Ng, and Xiaojun Wan, editors,
  {\em Proceedings of the 2019 Conference on Empirical Methods in Natural
  Language Processing and the 9th International Joint Conference on Natural
  Language Processing, {EMNLP-IJCNLP} 2019, Hong Kong, China, November 3-7,
  2019}, pages 1105--1116. Association for Computational Linguistics.

\bibitem[\protect\citename{Wang and Yang}2015]{DBLP:conf/emnlp/WangY15}
William~Yang Wang and Diyi Yang.
\newblock 2015.
\newblock That's so annoying!!!: {A} lexical and frame-semantic embedding based
  data augmentation approach to automatic categorization of annoying behaviors
  using {\#}petpeeve tweets.
\newblock In Llu{\'{\i}}s M{\`{a}}rquez, Chris Callison{-}Burch, Jian Su,
  Daniele Pighin, and Yuval Marton, editors, {\em Proceedings of the 2015
  Conference on Empirical Methods in Natural Language Processing, {EMNLP} 2015,
  Lisbon, Portugal, September 17-21, 2015}, pages 2557--2563. The Association
  for Computational Linguistics.

\bibitem[\protect\citename{Wei and Zou}2019]{DBLP:conf/emnlp/WeiZ19}
Jason~W. Wei and Kai Zou.
\newblock 2019.
\newblock {EDA:} easy data augmentation techniques for boosting performance on
  text classification tasks.
\newblock In Kentaro Inui, Jing Jiang, Vincent Ng, and Xiaojun Wan, editors,
  {\em Proceedings of the 2019 Conference on Empirical Methods in Natural
  Language Processing and the 9th International Joint Conference on Natural
  Language Processing, {EMNLP-IJCNLP} 2019, Hong Kong, China, November 3-7,
  2019}, pages 6381--6387. Association for Computational Linguistics.

\bibitem[\protect\citename{Wolf \bgroup et al.\egroup
  }2019]{wolf2019transformers}
Thomas Wolf, Lysandre Debut, Victor Sanh, Julien Chaumond, Clement Delangue,
  Anthony Moi, Pierric Cistac, Tim Rault, R{\'e}mi Louf, Morgan Funtowicz,
  et~al.
\newblock 2019.
\newblock Transformers: State-of-the-art natural language processing.
\newblock {\em arXiv preprint arXiv:1910.03771}.

\bibitem[\protect\citename{Yoo \bgroup et al.\egroup
  }2019]{DBLP:conf/aaai/YooSL19}
Kang~Min Yoo, Youhyun Shin, and Sang{-}goo Lee.
\newblock 2019.
\newblock Data augmentation for spoken language understanding via joint
  variational generation.
\newblock In {\em The Thirty-Third {AAAI} Conference on Artificial
  Intelligence, {AAAI} 2019, The Thirty-First Innovative Applications of
  Artificial Intelligence Conference, {IAAI} 2019, The Ninth {AAAI} Symposium
  on Educational Advances in Artificial Intelligence, {EAAI} 2019, Honolulu,
  Hawaii, USA, January 27 - February 1, 2019}, pages 7402--7409. {AAAI} Press.

\bibitem[\protect\citename{Zhao \bgroup et al.\egroup
  }2019a]{DBLP:conf/emnlp/ZhaoZY19}
Zijian Zhao, Su~Zhu, and Kai Yu.
\newblock 2019a.
\newblock Data augmentation with atomic templates for spoken language
  understanding.
\newblock In Kentaro Inui, Jing Jiang, Vincent Ng, and Xiaojun Wan, editors,
  {\em Proceedings of the 2019 Conference on Empirical Methods in Natural
  Language Processing and the 9th International Joint Conference on Natural
  Language Processing, {EMNLP-IJCNLP} 2019, Hong Kong, China, November 3-7,
  2019}, pages 3635--3641. Association for Computational Linguistics.

\bibitem[\protect\citename{Zhao \bgroup et al.\egroup
  }2019b]{zhao-etal-2019-data}
Zijian Zhao, Su~Zhu, and Kai Yu.
\newblock 2019b.
\newblock Data augmentation with atomic templates for spoken language
  understanding.
\newblock In {\em Proceedings of the 2019 Conference on Empirical Methods in
  Natural Language Processing and the 9th International Joint Conference on
  Natural Language Processing (EMNLP-IJCNLP)}, pages 3637--3643, Hong Kong,
  China, November. Association for Computational Linguistics.

\bibitem[\protect\citename{Zhong \bgroup et al.\egroup }2020]{zhong2020random}
Zhun Zhong, Liang Zheng, Guoliang Kang, Shaozi Li, and Yi~Yang.
\newblock 2020.
\newblock Random erasing data augmentation.
\newblock In {\em AAAI}, pages 13001--13008.

\end{thebibliography}
\bibliographystyle{acl}
\clearpage
\appendix

\section*{Appendix A. Hyperparameters}
\label{sec:appendix}
\begin{table}[!ht]
\centering
    \small
    \begin{tabular}{ll}
    \toprule
    Hyperparameter & Value  \\
    \toprule
    Learning rate & $10^{-5}$\\
    Dropout       &    0.1\\
    Mini-batch size      &   16 \\
    Optimizer      &   BertAdam \\
    Number of epoch      &   30 (\texttt{bert-base-uncased})  \\
    &   10 (\texttt{bert-large}, \\
    & \texttt{roberta-large},   \texttt{albert-xxl})\\
    Early stopping      &   10 \\
    \midrule
    $nb_{aug}$ & Tuned on \{2, 5, 10\}\\
    Nucleus sampling & top-\textit{p} = 0.9\\
    Max rotation & 3\\
    Max crop & 3\\
    \bottomrule

    \end{tabular}
    \caption{Hyperparameters used for the Transformer based models and data augmentation methods}
\label{tab:hyperparameter}
\end{table}

\end{document}